\documentclass[]{interact}

\usepackage{epstopdf}% To incorporate .eps illustrations using PDFLaTeX, etc.
\usepackage{subfigure}% Support for small, `sub' figures and tables
\usepackage{tabularx}
\usepackage{graphicx}
\usepackage{amsmath}
\usepackage{amssymb}
\usepackage{booktabs}
\usepackage{placeins} 
\usepackage{float} 
\usepackage{enumitem}
\usepackage{tabularx} % 引入 tabularx 宏包
\usepackage{longtable} % 引入 longtable 包
\usepackage{array} % 用于支持列类型定义
\usepackage{ragged2e} % 用于支持 \RaggedRight
\usepackage{booktabs} % 用于 \toprule, \midrule, \bottomrule
\usepackage{placeins}
\usepackage{afterpage} % 在导言区添加
\usepackage{caption} % 在导言区添加
\usepackage{booktabs} % 用于绘制表格的顶部、中部和底部线
\usepackage{adjustbox} % 用于调整表格宽度
\usepackage{graphicx} % 用于调整表格缩放
\usepackage{tabularx}

\theoremstyle{plain}% Theorem-like structures provided by amsthm.sty

\theoremstyle{definition}

\theoremstyle{remark}

\begin{document}

% \articletype{RESEARCH ARTICLE}

\title{MVNet: Hyperspectral Remote Sensing Image Classification Based on Hybrid Mamba-Transformer Vision Backbone Architecture}

\author{
\name{Guandong Li\textsuperscript{a}\thanks{CONTACT Guandong Li. Email: leeguandon@gmail.com} and Mengxia Ye\textsuperscript{b}}
\affil{\textsuperscript{a}iFLYTEK, Shushan, Hefei, Anhui, China; \textsuperscript{b}Aegon THTF,Qinghuai,Nanjing,Jiangsu,China}
}

\maketitle

\begin{abstract}
Hyperspectral image (HSI) classification faces challenges such as high-dimensional data, limited training samples, and spectral redundancy, which often lead to overfitting and insufficient generalization capability. This paper proposes a novel MVNet network architecture that integrates 3D-CNN's local feature extraction, Transformer's global modeling, and Mamba's linear complexity sequence modeling capabilities, achieving efficient spatial-spectral feature extraction and fusion. MVNet features a redesigned dual-branch Mamba module, including a State Space Model (SSM) branch and a non-SSM branch employing 1D convolution with SiLU activation, enhancing modeling of both short-range and long-range dependencies while reducing computational latency in traditional Mamba. The optimized HSI-MambaVision Mixer module overcomes the unidirectional limitation of causal convolution, capturing bidirectional spatial-spectral dependencies in a single forward pass through decoupled attention that focuses on high-value features, alleviating parameter redundancy and the curse of dimensionality. On IN, UP, and KSC datasets, MVNet outperforms mainstream hyperspectral image classification methods in both classification accuracy and computational efficiency, demonstrating robust capability in processing complex HSI data.
\end{abstract}

\begin{keywords}
Hyperspectral image classification; 3D convolution; Mamba; Transformer; Attention decoupling
\end{keywords}

\section{Introduction}

Hyperspectral remote sensing images (HSI) play a crucial role in spatial information applications due to their unique narrow-band imaging characteristics. The imaging equipment synchronously records both spectral and spatial position information of sampling points, integrating them into a three-dimensional data structure containing two-dimensional space and one-dimensional spectrum. As an important application of remote sensing technology, ground object classification demonstrates broad value in fields including transportation planning, ecological assessment, agricultural monitoring, land management, and geological surveys \cite{chang2003hyperspectral,bing2011intelligent}. However, hyperspectral remote sensing images face several challenges in ground object classification. First, hyperspectral data typically exhibits high-dimensional characteristics, with each pixel containing hundreds or even thousands of bands, leading to data redundancy and computational complexity while potentially causing the "curse of dimensionality," making classification models prone to overfitting under sparse sample conditions. Second, training samples are often limited, particularly for certain rare categories where annotation costs are high and distributions are imbalanced, further constraining model generalization capability. Additionally, ground object sparsity may also lead to insufficient spatial context information, causing loss in classification and recognition.

Deep learning methods for HSI classification \cite{zhang2019multi,zhang2019three,li2020hyperspectral,li2019doubleconvpool} have achieved significant progress. In \cite{lee2017going} and \cite{zhao2016spectral}, Principal Component Analysis (PCA) was first applied to reduce the dimensionality of the entire hyperspectral data, followed by extracting spatial information from neighboring regions using 2D CNN. Methods like 2D-CNN \cite{makantasis2015deep,chen2016deep} require separate extraction of spatial and spectral features, failing to fully utilize joint spatial-spectral information and necessitating complex preprocessing. Wang et al. \cite{wang2018fast} proposed a Fast Dense Spectral-Spatial Convolutional Network (FDSSC) based on dense networks, constructing 1D-CNN and 3D-CNN dense blocks connected in series. FSKNet \cite{li2022faster} introduced a 3D-to-2D module and selective kernel mechanism, while 3D-SE-DenseNet \cite{li2020hyperspectral} incorporated the SE mechanism into 3D-CNN to correlate feature maps between different channels, activating effective information while suppressing ineffective information in feature maps. DGCNet \cite{li2025dgcnet} designed dynamic grouped convolution (DGC) on 3D convolution kernels, where DGC introduces small feature selectors for each group to dynamically determine which part of input channels to connect based on activations of all input channels. Multiple groups can capture different complementary visual/semantic features of input images, enabling CNNs to learn rich feature representations. Currently widely used methods such as DFAN \cite{zhang2020deep}, MSDN \cite{zhang2019multi}, 3D-DenseNet \cite{zhang2019three}, and 3D-SE-DenseNet \cite{li2020hyperspectral} employ operations like dense connections. While dense connections directly link each layer to all its preceding layers, enabling feature reuse, they introduce redundancy when subsequent layers do not require early features. Additionally, some Transformer-based methods \cite{hong2021spectralformer,he2021spatial} employ grouped spectral embedding and transformer encoder modules to model spectral representations, but these methods have obvious shortcomings - they treat spectral bands or spatial patches as tokens and encode all tokens, resulting in significant redundant computations. However, HSI data already contains substantial redundant information, and their accuracy often falls short compared to 3D-CNN-based methods while requiring greater computational complexity. To leverage the advantages of both CNN and Transformer, many studies have attempted to combine CNN and Transformer to utilize local and global feature information of HSI. Sun et al. \cite{sun2022spectral} proposed a Spectral-Spatial Feature Tokenization Transformer (SSFTT) network that extracts shallow features through 3D and 2D convolutional layers and uses Gaussian-weighted feature tokens to extract high-level semantic features in the transformer encoder. CMTNet \cite{guo2024cmtnet} includes a spectral-spatial feature extraction module for shallow feature capture, a dual-branch structure combining CNN and Transformer branches for local and global feature extraction, and a multi-output constraint module. However, these structures often lack sufficient granularity in spatial-spectral segmentation, resulting in low utilization of spatial-spectral information, and the quadratic complexity of Transformer structures relative to sequence length further exacerbates computational redundancy and high computational costs. Recently, Mamba \cite{gu2023mamba,zhang2024survey} proposed a novel state space model (SSM) that achieves linear time complexity and outperforms or matches Transformers in various language modeling tasks. The core contribution of Mamba is a novel selective mechanism that efficiently processes long sequences. SpectralMamba \cite{yao2024spectralmamba} simplifies but adequately models HSI data dynamics in spatial-spectral feature space and hidden state space, thereby mitigating effects caused by spectral variability and spectral confusion.

3D-CNN possesses the capability to sample simultaneously in both spatial and spectral dimensions, maintaining the spatial feature extraction ability of 2D convolution while ensuring effective spectral feature extraction. 3D-CNN can directly process high-dimensional data, eliminating the need for preliminary dimensionality reduction of hyperspectral images. However, the 3D-CNN paradigm has significant limitations - when simultaneously extracting spatial and spectral features, it may incorporate irrelevant or inefficient spatial-spectral combinations into computations. For instance, certain spatial features may be prominent in specific bands while appearing as noise or irrelevant information in other bands, yet 3D convolution still forcibly combines these low-value features. Computational and dimensional redundancy can easily trigger overfitting risks, further limiting model generalization capability. The self-attention mechanism in Transformers can directly compute dependency relationships between any two positions in the input sequence, capturing long-range spectral correlations (e.g., specific absorption peaks or reflection valleys across multiple bands) and non-local spatial context information, which is crucial for identifying ground objects that rely on global spectral patterns or large-scale spatial structures. However, the quadratic complexity of the attention mechanism relative to sequence length makes Transformers computationally expensive during training and deployment. Therefore, to better integrate the advantages of 3D-CNN and Transformer in hyperspectral image classification tasks while overcoming their respective limitations, we introduce Mamba and redesign a new hybrid architecture. Mamba is an architecture based on state space models, whose core advantage lies in its selective scanning mechanism (Selective Scanning Mechanism) that captures long-range dependencies with linear complexity, significantly reducing computational costs compared to Transformer's self-attention mechanism (quadratic complexity). This enables Mamba to efficiently model long-range spectral correlations and non-local spatial context in high-dimensional HSI data while maintaining lower computational overhead. Through this hybrid architecture, 3D-CNN's local feature extraction capability, Transformer's global modeling capability, and Mamba's efficient sequence modeling capability are organically combined. This architecture can more efficiently process high-dimensional data in hyperspectral classification tasks, reduce computational redundancy, improve model generalization capability, while maintaining modeling capability for complex spatial-spectral patterns.

This work addresses the challenge of effectively extracting and fusing spatial context with fine spectral information in hyperspectral image (HSI) classification by proposing a novel network architecture called MVNet. Mamba's autoregressive formulation limits the model's ability to capture global context in visual tasks, especially in hyperspectral image classification where features in spatial and spectral dimensions need parallel processing rather than strict sequence dependencies. To address this, we redesigned the Mamba module to better adapt to the non-sequential requirements of HSI tasks. We introduced a symmetric branch without state space models (SSM), which processes inputs through 1D convolution and Sigmoid Linear Unit (SiLU) activation functions, compensating for global spatial and spectral information that might be lost due to sequence constraints in the SSM branch. The outputs of both SSM and non-SSM branches are projected into a lower-dimensional embedding space (half of the original embedding dimension) to maintain computational efficiency, followed by concatenation and final linear layer processing. This dual-branch design significantly enhances the model's ability to capture both short-range and long-range dependencies, overcoming the limitations of Mamba-based models in global context modeling, which typically rely on bidirectional scanning or cross-scanning modules that often introduce high computational latency. Furthermore, we redesigned the causal convolution in the Mamba block. The unidirectionality of causal convolution limits its modeling of bidirectional spatial-spectral relationships, whereas in hyperspectral data, spatial features (such as ground object boundaries or textures) and spectral features (such as cross-band absorption peaks) often need simultaneous consideration. The introduction of optimized convolution eliminates this directional constraint, enabling the HSI-MambaVision Mixer to capture bidirectional spatial-spectral dependencies in a single forward pass, thereby more efficiently modeling complex interactions in HSI data. This structure not only enhances representation of key spatial-spectral features through decoupled attention but also, compared to traditional 3D-CNN, can more effectively focus on high-value information while reducing redundant interference, avoiding uniform treatment of all spectral dimensions, thereby effectively alleviating parameter redundancy and the curse of dimensionality, ultimately improving classification accuracy.

The main contributions of this paper are as follows:

1. We propose a novel MVNet network architecture that integrates the advantages of 3D-CNN, Transformer, and Mamba. This paper designs a completely new MVNet network architecture for hyperspectral image (HSI) classification tasks. By organically combining 3D-CNN's local feature extraction capability, Transformer's global modeling capability, and Mamba's linear complexity sequence modeling capability, it achieves efficient spatial-spectral feature extraction and fusion. Through optimized hybrid architecture, MVNet significantly reduces computational redundancy and improves model generalization capability, achieving excellent classification accuracy on datasets such as Indian Pines and Pavia University.

2. We redesign the Mamba module to adapt to the non-sequential requirements of HSI classification. To address the limitations of Mamba's autoregressive properties in global context modeling for visual tasks, this paper introduces a dual-branch Mamba module, including a branch with state space model (SSM) and a symmetric branch without SSM. The non-SSM branch processes input through 1D convolution and SiLU activation function, compensating for global spatial and spectral information that might be lost due to sequence constraints in the SSM branch. The outputs of both branches are projected into low-dimensional embedding space and concatenated, effectively enhancing the model's ability to capture both short-range and long-range dependencies, overcoming the computational latency issues in traditional Mamba models.

3. We optimize causal convolution to capture bidirectional spatial-spectral dependencies. This paper proposes a novel HSI-MambaVision Mixer module that addresses the unidirectional limitation of causal convolution. By eliminating directional constraints, it captures bidirectional dependencies between spatial features (such as ground object boundaries and textures) and spectral features (such as cross-band absorption peaks) in a single forward pass. Compared to traditional 3D-CNN, this module enhances representation of key spatial-spectral features through decoupled attention mechanisms, effectively focusing on high-value information while alleviating issues of parameter redundancy and the curse of dimensionality, thereby significantly improving classification accuracy and computational efficiency.

\section{Hyperspectral Classification Method with Spatial-Spectral Transformer}
\subsection{Novel Structure Design in Hyperspectral Imaging}
\subsubsection{3D-CNN}

Given the sparse and finely clustered characteristics of hyperspectral ground objects, most methods consider how to enhance the model's feature extraction capability regarding spatial-spectral dimensions, making novel structure design an excellent approach to strengthen spatial-spectral dimension information representation. LGCNet \cite{li2025spatial} designed a learnable grouped convolution structure where both input channels and convolution kernel groups can be learned end-to-end through the network. DGCNet \cite{li2025dgcnet} designed dynamic grouped convolution, introducing small feature selectors for each group to dynamically determine which part of input channels to connect based on activations of all input channels. DACNet \cite{li2025efficient} designed dynamic attention convolution using SE to generate weights and multiple parallel convolutional kernels instead of single convolution. SG-DSCNet \cite{li2025spatialgeometryenhanced3ddynamic} designed a Spatial-Geometry Enhanced 3D Dynamic Snake Convolutional Neural Network, introducing deformable offsets in 3D convolution to increase kernel flexibility through constrained self-learning processes, thereby enhancing the network's regional perception of ground objects and proposing multiview feature fusion. WCNet \cite{li20253d} designed a 3D convolutional network integrating wavelet transform, introducing wavelet transform to expand the receptive field of convolution through wavelet convolution, guiding CNN to better respond to low frequencies through cascading. Each convolution focuses on different frequency bands of the input signal with gradually increasing scope. EKGNet \cite{li2025expert} includes a context-associated mapping network and dynamic kernel generation module, where the context-associated mapping module translates global contextual information of hyperspectral inputs into instructions for combining base convolutional kernels, while dynamic kernels are composed of K groups of base convolutions, analogous to K different types of experts specializing in fundamental patterns across various dimensions. KANet \cite{li2025dynamic} designed a more reasonable KAN convolution, embedding KAN's function learning capability into a 3D operator that maintains the locality of convolution operations and (a certain form of) parameter sharing characteristics, making it particularly suitable for processing three-dimensional data like hyperspectral images (two spatial dimensions + one spectral dimension). These methods significantly improve the spatial-spectral feature extraction and ground object perception capabilities of hyperspectral images by optimizing 3D convolution structures and introducing dynamic, flexible convolution kernel designs and wavelet transform techniques.

\subsubsection{Novel Designs in Transformer and Mamba}

Unlike these methods that focus on improving the convolution operation itself or introducing dynamic mechanisms, STNet \cite{li2025hyperspectral} adopts another strategy: introducing a carefully designed SpatioTemporalTransformer module within the 3D-CNN framework. This module aims to more effectively combine 3D-CNN's local feature extraction capability with Transformer's advantages in global dependency modeling and adaptive information integration through explicit spatial-spectral attention decoupling and dual gating mechanisms (adaptive fusion gating and gated feed-forward network) to address the unique challenges of HSI classification. MambaHSI \cite{li2024mambahsi} proposed Spatial-Spectral Mamba (SS-Mamba), which mainly consists of a spatial-spectral token generation module and multiple stacked spatial-spectral Mamba modules (SS-MB). First, the token generation module converts any given hyperspectral image cube into spatial and spectral token sequences. These tokens are then sent to stacked spatial-spectral Mamba modules. 3DSS-Mamba \cite{he20243dss} designed a Spectral-Spatial Token Generation (SSTG) module to transform HSI cubes into a series of 3D spectral-spatial tokens, introducing a three-dimensional spectral-spatial selective scanning (3DSS) mechanism that performs pixel-by-pixel selective scanning of 3D hyperspectral tokens along spectral and spatial dimensions. The 3DSS scanning mechanism combined with conventional mapping operations forms a three-dimensional spectral-spatial Mamba module (3DMB) capable of extracting global spectral-spatial semantic representations. However, the token generation and scanning path design are relatively complex, increasing the model's computational latency and parameter redundancy, limiting its efficiency in practical applications. The MVNet proposed in this paper overcomes the limitations of STNet, MambaHSI, and 3DSS-Mamba through innovative dual-branch Mamba modules and optimized HSI-MambaVision Mixer modules. MVNet combines SSM with non-SSM branches (based on 1D convolution and SiLU activation) to compensate for the shortcomings of traditional Mamba in global modeling while maintaining linear complexity. The newly designed HSI-MambaVision Mixer eliminates the unidirectional constraints of causal convolution, efficiently capturing bidirectional spatial-spectral dependencies and enhancing key feature representation. Compared to STNet's high computational complexity and the complex token generation of MambaHSI and 3DSS-Mamba, MVNet reduces redundancy through decoupled attention mechanisms, alleviating the curse of dimensionality.

\subsection{Design of Spatial-Spectral MambaVision}
Hyperspectral image sample data is scarce and exhibits sparse ground object characteristics, with uneven spatial distribution and substantial redundant information in the spectral dimension. Although 3D-CNN structures can utilize joint spatial-spectral information, how to more effectively achieve deep extraction of spatial-spectral information remains a noteworthy issue. As the core of convolutional neural networks, convolution kernels are generally regarded as information aggregators that combine spatial information and feature dimension information in local receptive fields. Convolutional neural networks consist of a series of convolutional layers, nonlinear layers, and downsampling layers, enabling them to capture image features from a global receptive field for image description. However, training a high-performance network is challenging, and much work has been done to improve network performance from the spatial dimension perspective. For example, the Residual structure achieves deep network extraction by fusing features produced by different blocks, while DenseNet enhances feature reuse through dense connections. 3D-CNN contains numerous redundant weights in feature extraction through convolution operations that simultaneously process spatial and spectral information of hyperspectral images. This redundancy is particularly prominent in joint spatial-spectral feature extraction: from the spatial dimension, ground objects in hyperspectral images are sparsely and unevenly distributed, and convolutional kernels may capture many irrelevant or low-information regions within local receptive fields; from the spectral dimension, hyperspectral data typically contains hundreds of bands with high correlation and redundancy between adjacent bands, making weight allocation of convolutional kernels along the spectral axis difficult to effectively focus on key features. Especially the redundant characteristics of spectral dimensions cause many convolutional parameters to only serve as "fillers" in high-dimensional data without fully mining deep patterns in joint spatial-spectral information. This weight redundancy not only increases computational complexity but may also weaken the model's representation capability for sparse ground objects and complex spectral features, thereby limiting 3D-CNN's performance in hyperspectral image processing.

MVNet proposes an innovative solution by embedding the HSI-MambaVision Mixer module and dual-branch Mamba module within the 3D-CNN architecture to efficiently address the challenges of high-dimensional redundancy and sparse ground object distribution in hyperspectral image (HSI) classification. The dual-branch Mamba structure includes an SSM branch that captures long-range spatial-spectral dependencies with linear complexity through selective scanning, and a non-SSM branch that compensates for missing global context through 1D convolution and SiLU activation. The outputs of both branches are projected into low-dimensional space and concatenated for fusion, reducing computational latency. The HSI-MambaVision Mixer optimizes non-causal convolution, eliminating unidirectional constraints to capture bidirectional spatial-spectral dependencies in a single forward pass, while focusing on key features through decoupled attention and suppressing redundancy. Compared to traditional 3D-CNN, MVNet combines Mamba's efficient sequence modeling with Transformer's global modeling for superior parameter utilization. Experiments show that MVNet outperforms traditional methods on Indian Pines and Pavia University datasets at lower computational costs, alleviating the curse of dimensionality and overfitting while demonstrating strong adaptability. MVNet marks a shift in hyperspectral classification towards flexible, efficient hybrid architectures, providing better performance and generalization for complex data processing.

\begin{figure}[h]
\centering
\includegraphics[width=0.5\linewidth]{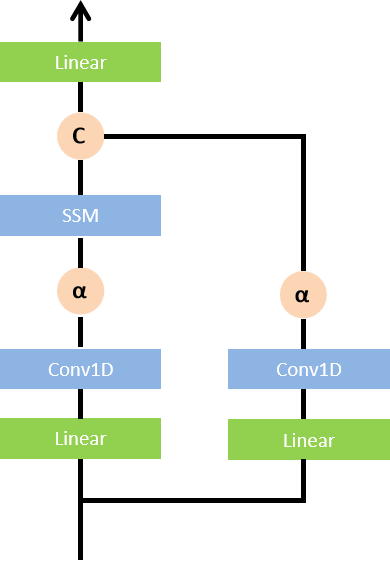}
\caption{Architecture of the MambaVision module}
\label{fig:mamba_module}
\end{figure}

\subsubsection{Mamba Preliminaries}
In Mamba, a 1D continuous input \( x(t) \in \mathbb{R} \) is transformed into \( y(t) \in \mathbb{R} \) through a learnable hidden state \( h(t) \in \mathbb{R}^M \), with parameters \( A \in \mathbb{R}^{M \times M} \), \( B \in \mathbb{R}^{M \times 1} \), and \( C \in \mathbb{R}^{1 \times M} \), as follows:
\begin{equation}
h'(t) = Ah(t) + Bx(t),
\end{equation}
\begin{equation}
y(t) = Ch(t),
\end{equation}

Discretization:
The continuous parameters \( A \), \( B \), and \( C \) are further converted into discrete parameters to improve computational efficiency~\cite{gu2021combining}. Specifically, given a time scale \( \Delta \), the zero-order hold rule can be applied to obtain discrete parameters \( \overline{A} \in \mathbb{R}^{M \times M} \), \( \overline{B} \in \mathbb{R}^{M \times 1} \), and \( \overline{C} \in \mathbb{R}^{1 \times M} \):
\begin{equation}
\bar{A} = \exp(\Delta A),
\end{equation}
\begin{equation}
\bar{B} = (\Delta A)^{-1}(\exp(\Delta A) - I) \cdot (\Delta B),
\end{equation}
\begin{equation}
\bar{C} = C,
\end{equation}

Equation 2 can be expressed with discrete parameters as:
\begin{equation}
h(t) = \bar{A}h(t - 1) + \bar{B}x(t),
\end{equation}
\begin{equation}
y(t) = \bar{C}h(t),
\end{equation}

Additionally, for an input sequence of size T, a global convolution with kernel K can be applied to compute the output of Eq. 4 as follows:
\begin{equation}
K = (CB, CAB, ..., CAT-1B),
\end{equation}
\begin{equation}
y = x * K,
\end{equation}

Selective Mamba further extends the S4 formulation by introducing a selection mechanism that enables input-dependent sequence processing. This allows the model's parameters \( B \), \( C \), and \( \Delta \) to be dynamically adjusted based on input and irrelevant information to be filtered out.

\subsubsection{Layer Architecture}
Given input \( X \in \mathbb{R}^{T \times C} \), where sequence length is \( T \) and embedding dimension is \( C \), the output of the \( n \)-th layer in stages 3 and 4 can be calculated as:
\begin{equation}
\hat{X}_n = \text{Mixer}(\text{Norm}(X_{n-1})) + X_{n-1},
\end{equation}
\begin{equation}
X_n = \text{MLP}(\text{Norm}(\hat{X}_n)) + \hat{X}_n,
\end{equation}

Norm and Mixer represent the selection of layer normalization (Layer Normalization) and token mixing blocks, respectively. For generality, Norm uses layer normalization. Given N layers, the first N/2 layers adopt MambaVision mixing blocks, while the remaining N/2 layers use self-attention. We describe the details of each mixer below.

\textbf{MambaVision Mixer:}
As shown in Figure 1, we redesigned the original Mamba mixer to make it more suitable for visual tasks. First, we recommend replacing causal convolution with regular convolution because causal convolution limits the direction of influence, which is unnecessary and restrictive for visual tasks. Additionally, we added a symmetric branch without SSM (sequential state model), consisting of an extra convolution and Sigmoid Linear Unit (SiLU) activation \cite{elfwing2018sigmoid}, to compensate for any content lost due to SSM's sequential constraints. Then, we concatenate the outputs of both branches and project them through a final linear layer. This combination ensures the final feature representation contains both sequential and spatial information, leveraging the advantages of both branches. We note that each branch's output is projected into an embedding space of size C/2 (i.e., half of the original embedding dimension) to maintain similar parameter counts as the original block design. Given input Xin, the output Xout of the MambaVision mixer is calculated as:
\begin{equation}
X_1 = \text{Scan}(\sigma(\text{Conv}(\text{Linear}(C, C/2)(X_{in})))),
\end{equation}
\begin{equation}
X_2 = \sigma(\text{Conv}(\text{Linear}(C, C/2)(X_{in}))),
\end{equation}
\begin{equation}
X_{out} = \text{Linear}(C/2, C)(\text{Concat}(X_1, X_2)),
\end{equation}

Here, \( \text{Linear}(C_{\text{in}}, C_{\text{out}})(\cdot) \) represents a linear layer with \( C_{\text{in}} \) and \( C_{\text{out}} \) as input and output embedding dimensions, respectively; \( \text{Scan} \) is the selective scanning operation described in~\cite{gu2023mamba}; \( \sigma \) is the activation function (here using SiLU); \( \text{Conv} \) and \( \text{Concat} \) represent 1D convolution and concatenation operations, respectively.

\textbf{Self-Attention:}
We use the generic multi-head self-attention mechanism:
\begin{equation}
\text{Attention}(Q, K, V) = \text{Softmax}(QK^T/\sqrt{d_h})V,
\end{equation}
where Q, K, V represent query, key, and value, respectively, and dh is the number of attention heads. Additionally, our framework allows attention computation in a windowed manner, similar to previous work \cite{liu2021swin,liu2022swin}.

\subsection{3D-CNN Framework for Hyperspectral Image Feature Extraction and Model Implementation}
To address challenges in hyperspectral remote sensing image (HSI) classification such as high-dimensional redundancy, sparse ground objects, and limited training samples, MVNet proposes an innovative 3D-CNN feature extraction framework that integrates 3D-CNN's local feature extraction capability, Transformer's global modeling capability, and Mamba's linear complexity sequence modeling capability for efficient spatial-spectral feature extraction and fusion. Through optimized dual-branch Mamba modules and HSI-MambaVision Mixer modules, MVNet significantly enhances modeling capability for complex spatial-spectral patterns while reducing computational overhead and parameter redundancy. Below are detailed descriptions of MVNet's feature extraction framework and model implementation.

\begin{figure}[h]
\centering
\includegraphics[width=0.9\linewidth]{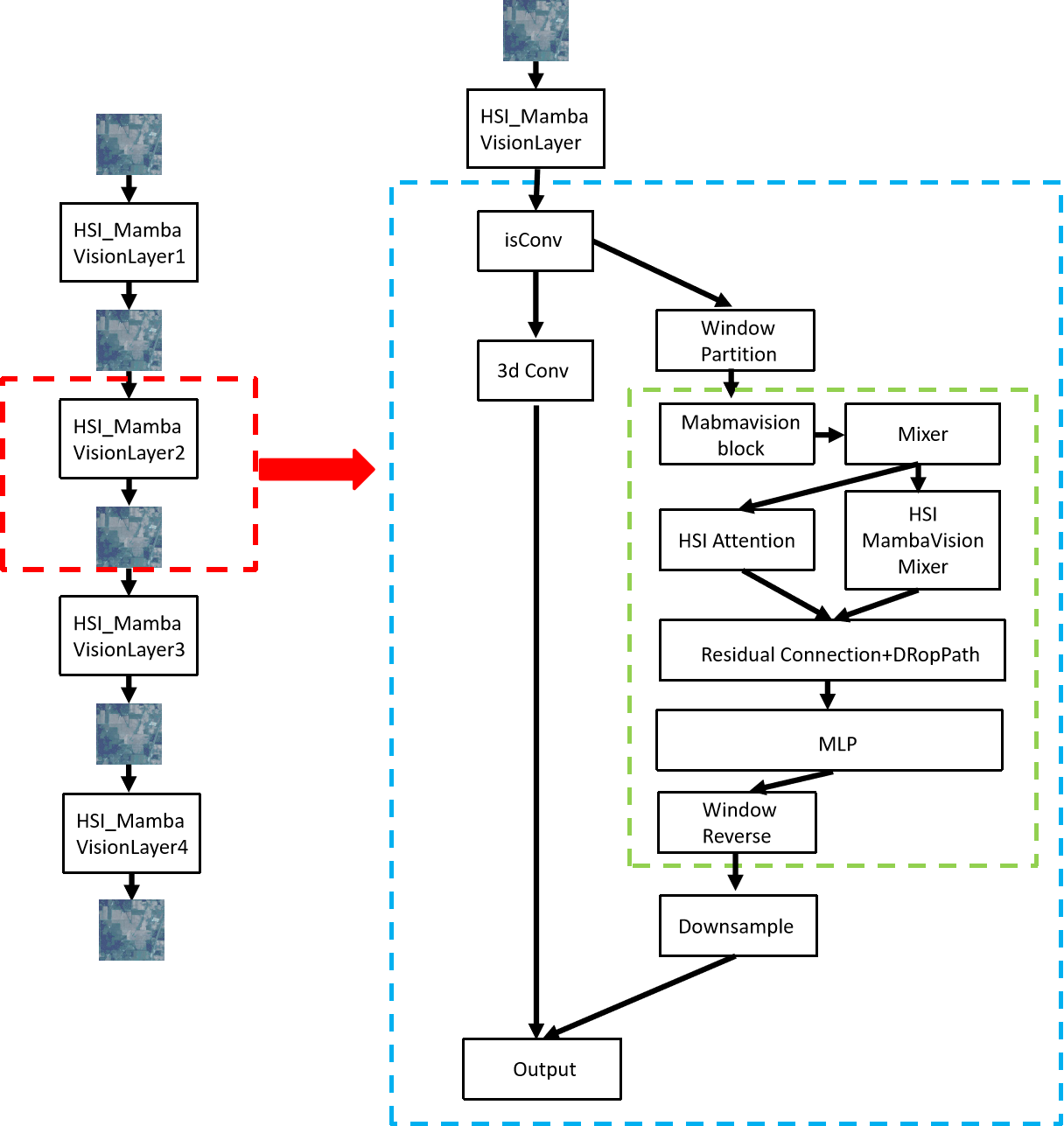}
\caption{Detailed design of HSI-MambaVision layer and overall hierarchical structure}
\label{fig:hsi_layer}
\end{figure}

\subsubsection{Spatial-Spectral Feature Extraction with 3D-CNN Backbone}
3D-CNN has become a mainstream method for hyperspectral image classification due to its ability to simultaneously extract features in both spatial and spectral dimensions. However, traditional 3D-CNN processing of high-dimensional HSI data often leads to overfitting and insufficient generalization capability due to redundant weights and forced fusion of low-value features. MVNet introduces an optimized feature extraction strategy in the 3D-CNN backbone network, enhancing focus on key spatial-spectral features through adaptive convolution kernel design and channel attention mechanisms.

Specifically, MVNet's 3D-CNN backbone network consists of multiple 3D convolutional layers, each extracting local features in both spatial and spectral dimensions through 3×3×3 convolution kernels. To avoid traditional 3D-CNN's uniform treatment of all bands, MVNet incorporates a channel attention module (Channel Attention Module) after each convolutional layer, weighting feature maps of different bands to prioritize activation of spectral features (such as specific absorption peaks or reflection valleys) that contribute more to classification tasks. Additionally, to further reduce redundancy, MVNet adopts a sparse convolution (Sparse Convolution) strategy, filtering out low-information feature regions through dynamic skip connections while retaining only spatial-spectral information meaningful for ground object classification.

\subsubsection{Integration of HSI-MambaVision Mixer Module}
One of MVNet's core innovations is the HSI-MambaVision Mixer module, which overcomes the unidirectional limitation of traditional Mamba's causal convolution through optimized non-causal convolution and dual-branch Mamba structure, enabling single-forward-pass capture of bidirectional dependencies in both spatial and spectral dimensions. The HSI-MambaVision Mixer module design is inspired by Mamba's selective scanning mechanism (Selective Scanning Mechanism) but has been redesigned for HSI classification's non-sequential requirements to more efficiently process high-dimensional data.

In implementation, the HSI-MambaVision Mixer module first divides the input hyperspectral data cube into spatial-spectral tokens, then processes them through a dual-branch Mamba structure. The dual-branch structure consists of:

\textbf{SSM Branch:} Utilizes Mamba's selective scanning mechanism to capture long-range spatial-spectral dependencies with linear complexity, particularly suitable for modeling cross-band spectral correlations and non-local spatial context information.

\textbf{Non-SSM Branch:} Processes input through 1D convolution and Sigmoid Linear Unit (SiLU) activation function, compensating for global spatial and spectral information that might be lost due to sequence constraints in the SSM branch.

Outputs from both branches are projected into a low-dimensional embedding space (half of the original embedding dimension), fused through concatenation operation, and generate comprehensive feature representations through a final linear layer. This dual-branch design not only preserves Mamba's linear complexity advantage but also significantly enhances the model's capability to model both short-range and long-range dependencies. Compared to traditional Transformer's quadratic complexity, the HSI-MambaVision Mixer module maintains higher classification accuracy.

\subsubsection{Cross-Dimensional Feature Fusion with Decoupled Attention}
To further improve spatial-spectral feature fusion efficiency, MVNet introduces a decoupled attention mechanism (Decoupled Attention Mechanism) between the 3D-CNN and HSI-MambaVision Mixer modules, separately weighting features in spatial and spectral dimensions. Traditional 3D-CNN and Transformer methods often suffer from computational redundancy or loss of key features due to dimensional coupling when fusing spatial-spectral features. MVNet's decoupled attention mechanism separately models spatial features (such as ground object boundaries and textures) and spectral features (such as cross-band absorption peaks), then enhances representation of high-value features through weighted fusion.

Specifically, the decoupled attention mechanism consists of two submodules:

\textbf{Spatial Attention Submodule:} Computes attention weights for ground object boundaries and textures based on spatial dimension feature maps, enhancing perception capability for sparse ground objects.

\textbf{Spectral Attention Submodule:} Computes cross-band correlation weights based on spectral dimension feature maps, prioritizing focus on bands that contribute more to classification tasks.

Outputs from both submodules are fused through weighted concatenation and linear projection to form comprehensive spatial-spectral feature representations, significantly alleviating the curse of dimensionality.

\begin{figure}[h]
\centering
\includegraphics[width=0.9\linewidth]{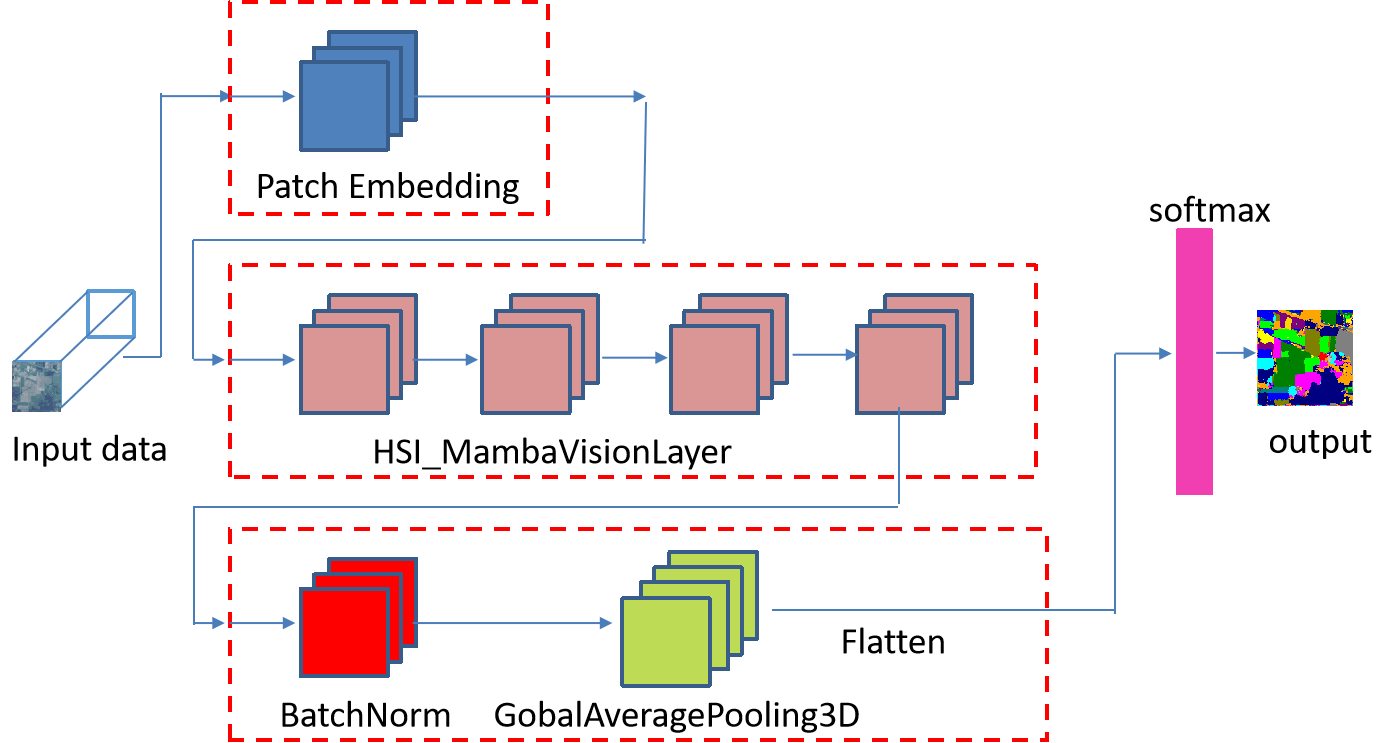}
\caption{Overall network architecture diagram}
\label{fig:overall_arch}
\end{figure}

\section{Experiments and Analysis}
To evaluate the performance of MVNet, we conducted experiments on three representative hyperspectral datasets: Indian Pines, Pavia University, and Kennedy Space Center (KSC). The classification metrics include Overall Accuracy (OA), Average Accuracy (AA), and Kappa coefficient.

\subsection{Experimental Datasets}
\subsubsection{Indian Pines Dataset}
\begin{figure}[h]
\centering
\includegraphics[width=0.9\linewidth]{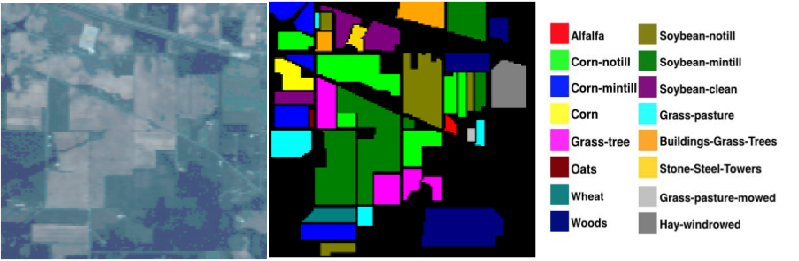}
\caption{False color composite and ground truth labels of Indian Pines dataset}
\label{fig:indian_pines}
\end{figure}

The Indian Pines dataset was collected in June 1992 by the AVIRIS (Airborne Visible/Infrared Imaging Spectrometer) sensor over a pine forest test site in northwestern Indiana, USA. The dataset consists of $145\times145$ pixel images with a spatial resolution of 20 meters, containing 220 spectral bands covering the wavelength range of 0.4--2.5$\mu$m. In our experiments, we excluded 20 bands affected by water vapor absorption and low signal-to-noise ratio (SNR), utilizing the remaining 200 bands for analysis. The dataset encompasses 16 land cover categories including grasslands, buildings, and various crop types. Figure~\ref{fig:indian_pines} displays the false-color composite image and spatial distribution of ground truth samples.

\subsubsection{Pavia University Dataset}
\begin{figure}[h]
\centering
\includegraphics[width=0.7\linewidth]{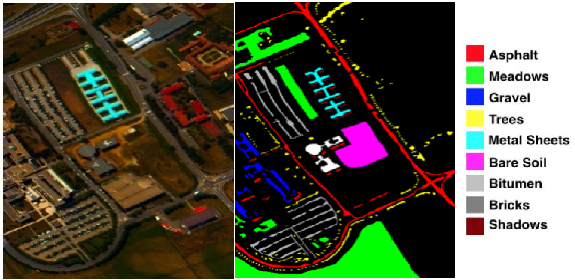}
\caption{False color composite and ground truth labels of Pavia University dataset}
\label{fig:pavia_university}
\end{figure}

The Pavia University dataset was acquired in 2001 by the ROSIS imaging spectrometer over the Pavia region in northern Italy. The dataset contains images of size $610\times340$ pixels with a spatial resolution of 1.3 meters, comprising 115 spectral bands in the wavelength range of 0.43--0.86$\mu$m. For our experiments, we removed 12 bands containing strong noise and water vapor absorption, retaining 103 bands for analysis. The dataset includes 9 land cover categories such as roads, trees, and roofs. Figure~\ref{fig:pavia_university} shows the spatial distribution of different classes.

\subsubsection{Kennedy Space Center Dataset}
The KSC dataset was collected on March 23, 1996 by the AVIRIS imaging spectrometer over the Kennedy Space Center in Florida. The AVIRIS sensor captured 224 spectral bands with 10 nm width, centered at wavelengths from 400 to 2500 nm. Acquired from an altitude of approximately 20 km, the dataset has a spatial resolution of 18 meters. After removing bands affected by water absorption and low signal-to-noise ratio (SNR), we used 176 bands for analysis, which represent 13 defined land cover categories.

\begin{table}[h]
\begin{minipage}{\textwidth}
\centering
\makeatletter
\def\@makecaption#1#2{%
    \vskip\abovecaptionskip
    \centering % 确保标题居中
    \small #1: #2\par
    \vskip\belowcaptionskip
}
\makeatother
\caption{OA, AA and Kappa metrics for different training set ratios on the Indian Pines dataset}
\begin{adjustbox}{width=0.5\columnwidth}
\begin{tabular}{cccc}
\toprule
Training Ratio & OA & AA & Kappa \\
\midrule
2:1:7 & 93.26 & 87.39 & 92.31 \\
3:1:6 & 97.07 & 94.68 & 96.66 \\
4:1:5 & 98.96 & 99.29 & 98.82 \\
5:1:4 & 99.29 & 99.25 & 99.19 \\
6:1:3 & 99.64 & 99.63 & 99.59 \\
\bottomrule
\end{tabular}
\label{tab:indian_ratios}
\end{adjustbox}
% \end{table}
    \end{minipage}

 \vspace*{10pt} % 可选：减少间距

\begin{minipage}{\textwidth}
% \begin{table}[t]
\centering
\makeatletter
\def\@makecaption#1#2{%
    \vskip\abovecaptionskip
    \centering % 确保标题居中
    \small #1: #2\par
    \vskip\belowcaptionskip
}
\makeatother
\caption{OA, AA and Kappa indicators for different training set ratios on the Pavia University dataset}
\begin{tabular}{cccc}
\toprule
Training Ratio & OA & AA & Kappa \\
\midrule
2:1:7 & 99.54 & 99.50 & 99.39 \\
3:1:6 & 99.81 & 99.73 & 99.74 \\
4:1:5 & 99.44 & 99.32 & 99.26 \\
5:1:4 & 99.94 & 99.92 & 99.92 \\
6:1:3 & 99.96 & 99.96 & 99.95 \\
\bottomrule
\end{tabular}
\label{tab:pavia_ratios}
% \end{table}
        \end{minipage}

 \vspace*{10pt} % 可选：减少间距

\begin{minipage}{\textwidth}        
% \begin{table}[t]
\centering
\makeatletter
\def\@makecaption#1#2{%
    \vskip\abovecaptionskip
    \centering % 确保标题居中
    \small #1: #2\par
    \vskip\belowcaptionskip
}
\makeatother
\caption{OA, AA and Kappa metrics for different training set proportions on the KSC dataset}
\begin{tabular}{cccc}
\toprule
Training Ratio & OA & AA & Kappa \\
\midrule
2:1:7 & 96.32 & 94.99 & 95.91 \\
3:1:6 & 98.85 & 98.41 & 98.72 \\
4:1:5 & 99.42 & 98.95 & 99.36 \\
5:1:4 & 99.76 & 99.59 & 99.73 \\
6:1:3 & 99.67 & 99.65 & 99.64 \\
\bottomrule
\end{tabular}
\label{tab:ksc_ratios}
        \end{minipage}
\end{table}

\subsection{Experimental Analysis}
MVNet was trained for 80 epochs on all three datasets using the Adam optimizer. The experiments were conducted on a platform with four 80GB A100 GPUs. For analysis, we employed the MVNet architecture with four stages, where each stage contained 1, 3, 8, and 16 HSI-MambaVisionLayers respectively. The window sizes were set to 4, 4, 7, 7, heads to 2, 4, 8, 16, mlp ratio to 4, and drop rate to 0.2. The conv3d dimension in Patchembed was set to 80.

\subsubsection{Data Partitioning Ratio}
For hyperspectral data with limited samples, the training set ratio significantly impacts model performance. To systematically evaluate the sensitivity of data partitioning strategies, we compared the model's generalization performance under different Train/Validation/Test ratios. Experiments show that with limited training samples, a 6:1:3 ratio effectively balances learning capability and evaluation reliability on Indian Pines - this configuration allocates 60\% samples for training, 10\% for validation (enabling early stopping to prevent overfitting), and 30\% for statistically significant testing. MVNet adopted 6:1:3 ratio on Pavia University and 5:1:4 ratio on KSC datasets, with 11×11 neighboring pixel blocks to balance local feature extraction and spatial context integrity.

\subsubsection{Neighboring Pixel Blocks}
The network performs edge padding on the input $145\times145\times103$ image (using Indian Pines as an example), transforming it into a $155\times155\times103$ image. On this padded image, it sequentially selects adjacent pixel blocks of size $M\times N\times L$, where $M\times N$ represents the spatial sampling size and $L$ is the full spectral dimension. Large original images are unfavorable for convolutional feature extraction, leading to slower processing speeds, temporary memory spikes, and higher hardware requirements. Therefore, we adopt adjacent pixel block processing. The block size is a crucial hyperparameter - too small blocks may result in insufficient receptive fields for convolutional feature extraction, leading to poor local performance. As shown in Tables 4-6, on the Indian Pines dataset, accuracy shows significant improvement when block size increases from 7 to 17. However, the accuracy gain diminishes with larger blocks, showing a clear threshold effect. At block size 17, accuracy even decreases slightly, a phenomenon also observed in Pavia University and KSC datasets. Consequently, we select block size 13 for Indian Pines and 17 for Pavia University and KSC datasets.

\begin{table}[h]
\centering
\caption{OA, AA and Kappa indicators under different adjacent pixel block sizes on the Indian pines dataset}
\begin{tabular}{cccc}
\toprule
Block Size (M=N) & OA & AA & Kappa \\
\midrule
7 & 99.22 & 99.25 & 99.11 \\
9 & 99.28 & 98.15 & 99.18 \\
11 & 99.64 & 99.63 & 99.59 \\
13 & 99.74 & 99.67 & 99.70 \\
15 & - & - & - \\
17 & - & - & - \\
\bottomrule
\end{tabular}
\label{tab:indian_blocks}
\end{table}

\begin{table}[h]
\centering
\caption{OA, AA and Kappa indicators under different adjacent pixel block sizes on the Pavia University dataset}
\begin{tabular}{cccc}
\toprule
Block Size (M=N) & OA & AA & Kappa \\
\midrule
7 & 99.78 & 99.70 & 99.71 \\
9 & 99.91 & 99.92 & 99.89 \\
11 & 99.96 & 99.96 & 99.95 \\
13 & 99.99 & 99.98 & 99.98 \\
15 & 99.99 & 99.99 & 99.99 \\
17 & 1 & 1 & 1 \\
\bottomrule
\end{tabular}
\label{tab:pavia_blocks}
\end{table}

\begin{table}[h]
\centering
\caption{OA, AA and Kappa indicators under different adjacent pixel block sizes under the KSC dataset}
\begin{tabular}{cccc}
\toprule
Block Size (M=N) & OA & AA & Kappa \\
\midrule
7 & 98.79 & 98.19 & 98.66 \\
9 & 99.42 & 98.96 & 99.36 \\
11 & 99.76 & 99.59 & 99.73 \\
13 & 99.86 & 99.66 & 99.84 \\
15 & 99.90 & 99.87 & 99.89 \\
17 & 1 & 1 & 1 \\
\bottomrule
\end{tabular}
\label{tab:ksc_blocks}
\end{table}

\subsection{Experimental Results}
On Indian Pines dataset, MVNet uses input size of $13\times13\times200$; on Pavia University dataset, $13\times13\times103$; and on KSC dataset, $17\times17\times176$. We compare MVNet-base with SSRN \cite{zhong2017spectral}, 3D-CNN \cite{zhao2016spectral}, 3D-SE-DenseNet, Spectralformer \cite{yang2022hyperspectral}, Hit \cite{li2024cmtnet}, and dgcnet. As shown in Tables 8-9, all three MVNet variants achieve leading accuracy. Figure 8 shows classification result maps. Figures 9-10 display training and validation loss and accuracy curves, demonstrating rapid convergence and stable accuracy improvement.

\begin{table*}[!ht]
\centering
\makeatletter
\def\@makecaption#1#2{%
    \vskip\abovecaptionskip
    \centering % 确保标题居中
    \small #1: #2\par
    \vskip\belowcaptionskip
}
\makeatother
\caption{Comparison of the classification accuracies (\%) of different methods for the Indian Pines dataset}
\resizebox{\textwidth}{!}{
\begin{tabular}{@{}lccccccc@{}}
\toprule
Class & SSRN & 3D-CNN & 3D-SE-DenseNet & DGCNet & Hit & Spectralformer & MVNet \\
\midrule
1 & 100 & 96.88 & 95.87 & 100 & Hit & 70.52 & 1 \\
2 & 99.85 & 98.02 & 98.82 & 99.47 & 94.25 & 81.89 & 1 \\
3 & 99.83 & 97.74 & 99.12 & 99.51 & 92.68 & 91.30 & 1 \\
4 & 100 & 96.89 & 94.83 & 97.65 & 78.55 & 95.53 & 1 \\
5 & 99.78 & 99.12 & 99.86 & 100 & 86.73 & 85.51 & 1 \\
6 & 99.81 & 99.41 & 99.33 & 99.88 & 85.33 & 99.32 & 1 \\
7 & 100 & 88.89 & 97.37 & 100 & 98.32 & 81.81 & 1 \\
8 & 100 & 100 & 100 & 100 & 92.00 & 75.48 & 1 \\
9 & 0 & 100 & 100 & 100 & 94.63 & 73.76 & 1 \\
10 & 100 & 100 & 99.48 & 98.85 & 64.86 & 98.77 & 99.34 \\
11 & 99.62 & 99.33 & 98.95 & 99.72 & 89.48 & 93.17 & 99.46 \\
12 & 99.17 & 97.67 & 95.75 & 99.56 & 94.40 & 78.48 & 99.45 \\
13 & 100 & 99.64 & 99.28 & 100 & 89.32 & 100 & 1 \\
14 & 98.87 & 99.65 & 99.55 & 99.87 & 99.46 & 79.49 & 1 \\
15 & 100 & 96.34 & 98.70 & 100 & 97.23 & 100 & 1 \\
16 & 98.51 & 97.92 & 96.51 & 98.30 & 68.71 & 100 & 86.43 \\
\midrule
OA & 99.62±0.00 & 98.23±0.12 & 98.84±0.18 & 99.58 & 91.67 & 81.76 & 99.74 \\
AA & 93.46±0.50 & 98.80±0.11 & 98.42±0.56 & 99.55 & 87.85 & 87.81 & 99.67 \\
K & 99.57±0.00 & 97.96±0.53 & 98.60±0.16 & 99.53 & 88.12 & 79.19 & 99.70 \\
\bottomrule
\end{tabular}
}
\label{tab:indian_results}
\end{table*}

\begin{figure*}[!ht]
\centering
\includegraphics[width=0.9\linewidth]{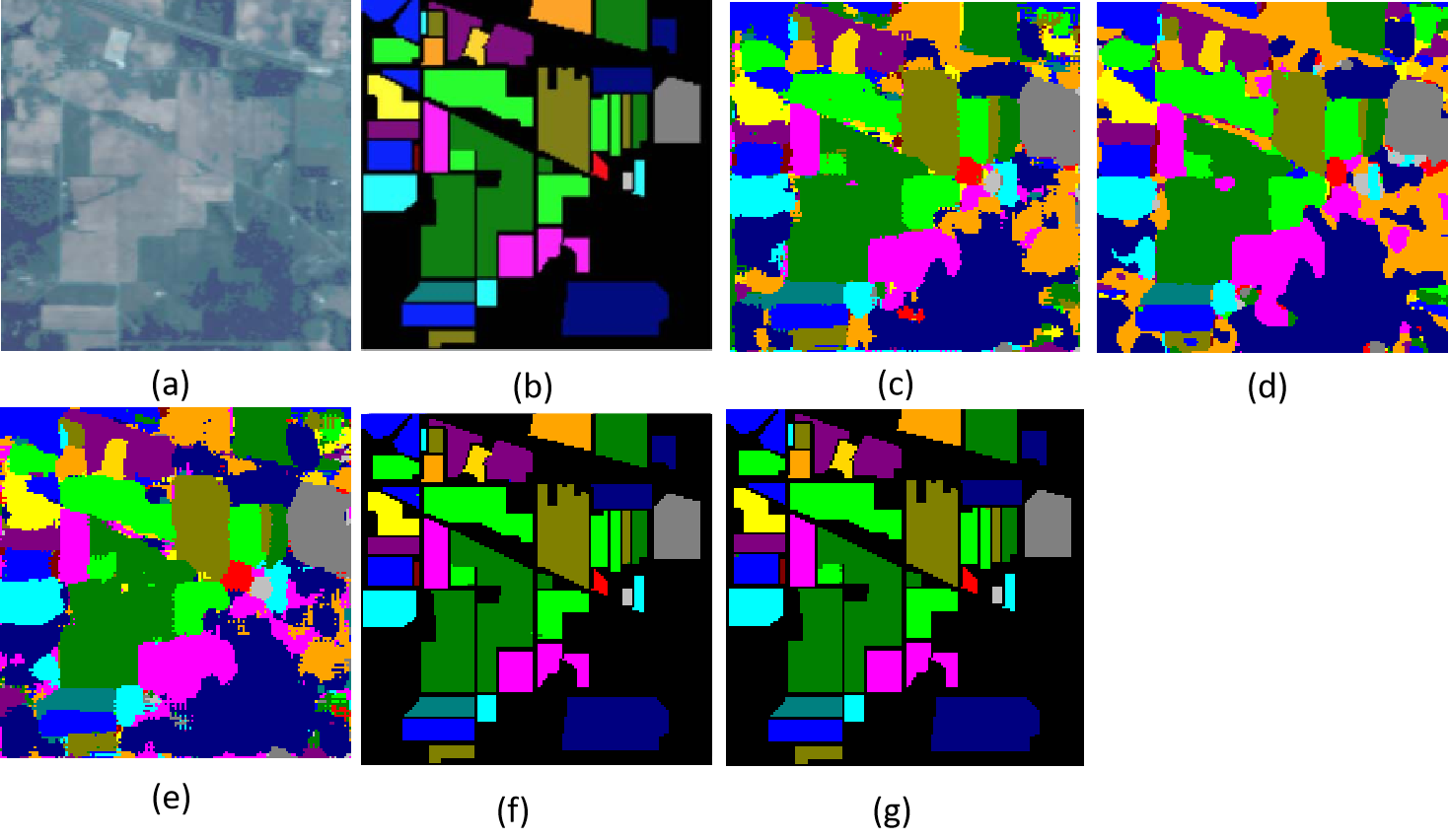}
\caption{Classification results of the best models for the Indian Pines dataset: (a) false color image, (b) ground-truth labels, (c)-(g) classification results of the SSRN, 3D-CNN, 3D-SE-DenseNet-BC, DGCNet, and MVNet}
\label{fig:classification}
\end{figure*}

\begin{figure*}[!ht]
\centering
\includegraphics[width=0.9\linewidth]{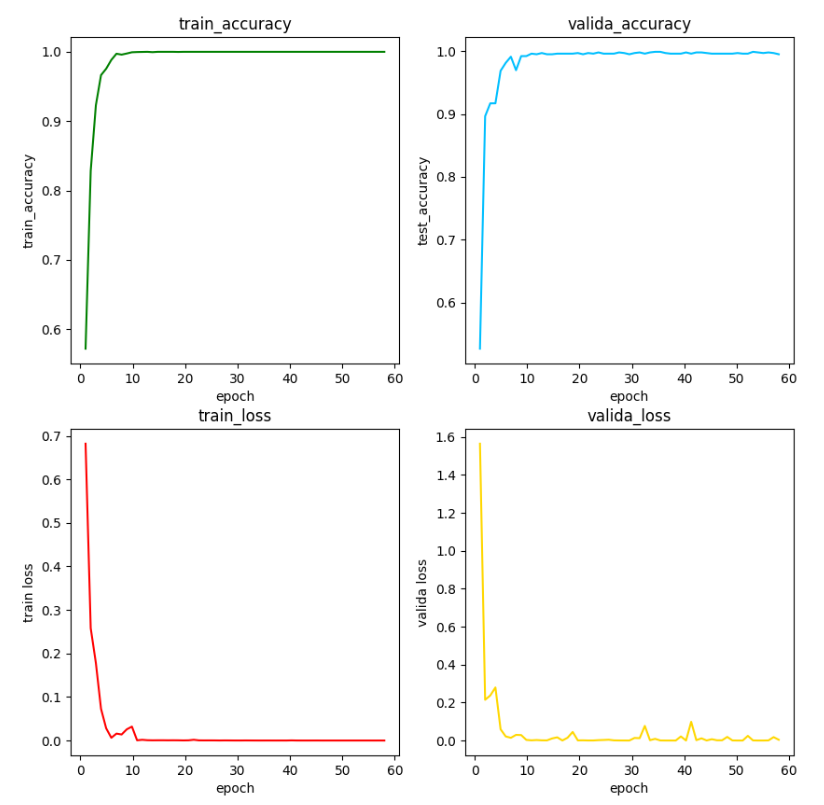}
\caption{Training and validation curves of MVNet on Indian Pines showing loss and accuracy evolution}
\label{fig:training_indian}
\end{figure*}

\begin{table*}[!ht]
\centering
\makeatletter
\def\@makecaption#1#2{%
    \vskip\abovecaptionskip
    \centering % 确保标题居中
    \small #1: #2\par
    \vskip\belowcaptionskip
}
\makeatother
\caption{Classification accuracy (\%) comparison on Pavia University dataset}
\resizebox{\textwidth}{!}{
\begin{tabular}{@{}lcccccc@{}}
\toprule
Class & SSRN & 3D-CNN & 3D-SE-DenseNet & Hit & Spectralformer & MVNet \\
\midrule
1 & 89.93 & 99.96 & 99.32 & 96.19 & 82.73 & 99.95 \\
2 & 86.48 & 99.99 & 99.87 & 92.79 & 94.03 & 1 \\
3 & 99.95 & 99.64 & 96.76 & 93.21 & 73.66 & 1 \\
4 & 95.78 & 99.83 & 99.23 & 97.33 & 93.75 & 1 \\
5 & 97.69 & 99.81 & 99.64 & 99.96 & 99.28 & 1 \\
6 & 95.44 & 99.98 & 99.80 & 99.91 & 90.75 & 1 \\
7 & 84.40 & 97.97 & 99.47 & 98.22 & 87.56 & 1 \\
8 & 100 & 99.56 & 99.32 & 99.15 & 95.81 & 99.91 \\
9 & 87.24 & 100 & 100 & 99.77 & 94.21 & 1 \\
\midrule
OA & 92.99±0.39 & 99.79±0.01 & 99.48±0.02 & 92.00 & 91.07 & 99.98 \\
AA & 87.21±0.25 & 99.75±0.15 & 99.16±0.37 & 93.24 & 90.20 & 99.98 \\
K & 90.58±0.18 & 99.87±0.27 & 99.31±0.03 & 89.77 & 88.05 & 99.98 \\
\bottomrule
\end{tabular}
}
\label{tab:pavia_results}
\end{table*}

\begin{figure*}[!ht]
\centering
\includegraphics[width=0.9\linewidth]{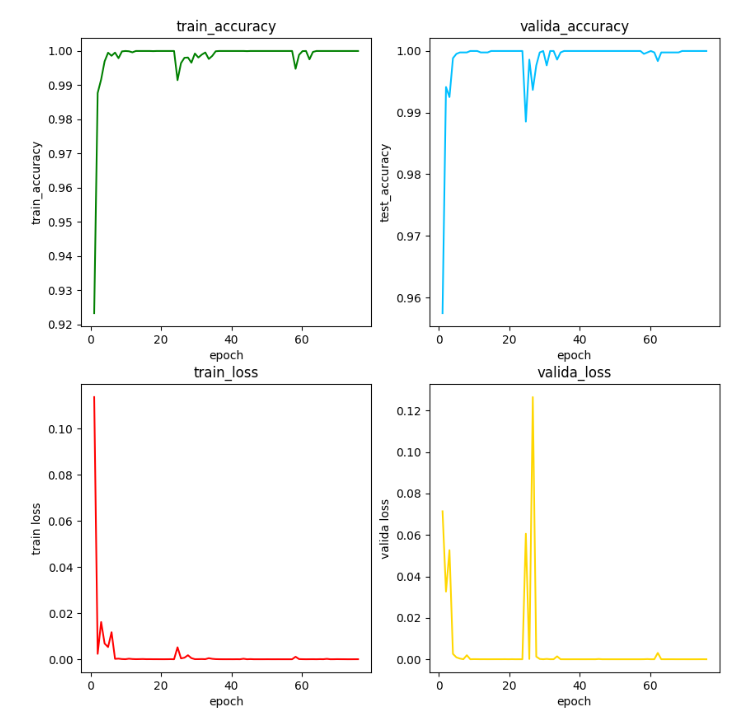}
\caption{Training and validation curves of MVNet on Pavia University showing loss and accuracy evolution}
\label{fig:training_pavia}
\end{figure*}

\section{Conclusion}
This paper proposes a novel MVNet architecture whose core innovation lies in its dual-branch Mamba module and optimized HSI-MambaVision Mixer module. The architecture integrates 3D-CNN's local feature extraction capability, Transformer's global modeling capability, and Mamba's linear complexity sequence modeling capability for efficient spatial-spectral feature extraction and fusion. The dual-branch Mamba module, through collaborative work between SSM and non-SSM branches (based on 1D convolution and SiLU activation), compensates for traditional Mamba's shortcomings in global context modeling while maintaining linear complexity. The newly designed HSI-MambaVision Mixer eliminates unidirectional constraints of causal convolution, efficiently capturing bidirectional spatial-spectral dependencies in a single forward pass while enhancing key feature representation through decoupled attention. Compared to traditional 3D-CNN, MVNet demonstrates powerful feature representation capability, particularly effective in reducing overfitting risks in small-sample and high-noise scenarios. Through decoupled attention mechanisms that focus on key features while suppressing redundant information, integrated within the 3D-CNN architecture to leverage its feature reuse advantages, MVNet provides an efficient and robust solution for hyperspectral image classification that successfully addresses challenges posed by sparse ground object distribution and spectral redundancy. Experiments show that MVNet outperforms traditional methods on Indian Pines and Pavia University datasets at lower computational costs, alleviating the curse of dimensionality and overfitting while demonstrating strong adaptability. MVNet marks a shift in hyperspectral classification towards flexible, efficient hybrid architectures, providing better performance and generalization for complex data processing.

\section*{Data Availability Statement}
The datasets used in this study are publicly available and widely used benchmark datasets in the hyperspectral image analysis community.

{\small
\bibliographystyle{template}
\bibliography{template}
}

\end{document}